# CultureBERT: Measuring Corporate Culture With Transformer-Based Language Models

Sebastian Koch*    Stefan Pasch§

This Version: June 19, 2023†

## Abstract

This paper introduces transformer-based language models to the literature measuring corporate culture from text documents. We compile a unique data set of employee reviews that were labeled by human evaluators with respect to the information the reviews reveal about the firms' corporate culture. Using this data set, we fine-tune state-of-the-art transformer-based language models to perform the same classification task. In out-of-sample predictions, our language models classify 17 to 30 percentage points more of employee reviews in line with human evaluators than traditional approaches of text classification. We make our models publicly available.

*Corresponding author. Faculty of Economics and Business, Goethe University Frankfurt, Theodor-W.-Adorno-Platz 4, 60323 Frankfurt am Main, Germany. Email: s.koch@its.uni-frankfurt.de. Telephone: +49 (69) 798-34825. Telefax: +49 (69) 798-35021.

§stefan.pasch@outlook.com

# 1. Introduction

In recent years, significant advancements in computational linguistics have allowed management scholars to better understand concepts that are hard to quantify, such as innovation (Bellstam *et al.*, 2021), market orientation (Andreou *et al.*, 2020), or corporate culture (Srivastava *et al.*, 2018). With respect to corporate culture in particular, researchers have increasingly tried to measure it by analyzing large textual data sets, consisting of, for example, employee reviews (Corritore *et al.*, 2020; Grennan, 2019; Pasch, 2018), annual reports (Andreou *et al.*, 2022; Fiordelisi and Ricci, 2014; Nguyen *et al.*, 2019; Wang *et al.*, 2021), email messages (Srivastava *et al.*, 2018), or earnings calls (Li *et al.*, 2021). However, studies measuring corporate culture by applying computational linguistics exhibit two major weaknesses. First, most computational linguistic techniques do not take the semantic context of words into account. Second, most studies do not evaluate in how far their measures of corporate culture align with the assessment of human evaluators. In addition to these weaknesses, most studies do not publish their language models in a way that allows other researchers to easily benchmark them against other computational linguistic techniques. In this paper, we address all of these issues.

We construct a unique data set of employee reviews that were labeled by human evaluators with respect to the information the reviews reveal about the firm's corporate culture. This labeled data set allows us to apply *supervised* machine learning to measure corporate culture. More specifically, we fine-tune state-of-the-art transformer-based language models and show that they further improve on the computational linguistic techniques applied in the literature to measure corporate culture so far. Transformer-based language models allow to capture complex aspects of textual information by taking the surrounding context into consideration. Their development has accelerated since the publication of *Bidirectional Encoder Representations from Transformers* (BERT; Devlin *et al.*, 2018). BERT set new high scores in various natural language processing tasks, such as question-answering, fact-checking, and text classification (González-Carvajal and Garrido-Merchán, 2020). While transformer-based language models have already been used in financial sentiment analysis (Araci, 2019), patent classification (Lee and Hsiang, 2020), or biomedical text mining (Lee *et al.*, 2020), we are not aware of any application with respect to corporate culture.

Our procedure of fine-tuning transformer-based language models for corporate culture classification is as follows:

- We randomly draw a sample of 2,000 employee reviews from a leading career community website.
- We create a labeled data set by determining for each of the 2,000 employee reviews whether it contains information either in accordance with or in opposition to the four culture dimensions of the Competing Values Framework (CVF; Quinn and Rohrbaugh, 1983). Moreover, we assign each review to the CVF's culture dimension that best matches the employee's description.
- We use a subset of this labeled data set to fine-tune transformer-based language models to classify employee reviews in the same way.

We find that our language models for corporate culture are more accurate than traditional computational linguistic techniques in classifying employee reviews with respect to corporate culture in the same way as human evaluators. Specifically, our models achieve a 17 to 30 percentage point higher accuracy than the dictionary method — a particularly popular approach of measuring corporate culture by comparing the text document of interest with a set of words that are thought to be characteristic of the cultural dimension under study. Moreover, our language models outperform other machine learning-based text classifiers by three to fifteen percentage points in accuracy. Our findings illustrate the usefulness of transformer-based language models not only for measuring corporate culture but also for studying organizational phenomena more generally.

Different from most other studies, we make our language models publicly available.[1] This allows other researchers to measure corporate culture from employee reviews or similar text documents without having to build their own language model for corporate culture classification. Moreover, it allows other researchers to benchmark our transformer-based language models against alternative computational linguistic techniques.

The remainder of the paper is organized as follows. In the following section, we explain how our paper relates to the literature. In section 3, we present our data. In section 4, we explain how we constructed our language models for corporate culture classification. In section 5, we compare our models' performance in classifying employee reviews with the performance of traditional computational linguistic techniques. In the last section, we discuss our findings.

## 2. Related Literature

### 2.1 Competing Values Framework

We follow O'Reilly and Chatman (1996, p. 160) and define corporate culture as "a system of shared values (that define what is important) and norms that define appropriate attitudes and behaviors for organizational members (how to feel and behave)." To create a labeled data set of employee reviews, we need a framework that lets us decide how to group the values, norms, and behaviors expressed in employee reviews into specific types of corporate culture. We make use of the Competing Values Framework (CVF; Quinn and Rohrbaugh, 1983), which is one of the frameworks most widely used for measuring corporate culture in both scientific research and practice (Hartnell *et al.*, 2011; Chatman and O'Reilly, 2016).

According to the CVF, organizations can be described along two basic value dimensions: their *focus* and their preference for *structure*. With respect to the focus dimension, organizations can be placed on a continuum between an internal focus, with an emphasis on internal capabilities and integration, and an external focus, with an emphasis on external opportunities and differentiation from competitors (Hartnell *et al.*, 2019). With respect to the structure dimension, organizations can be distinguished based on whether they prefer stability and control or flexibility and change. Plotting the focus and structure dimensions in a two-dimensional space yields four quadrants, of which each stands for one type of corporate culture: clan, hierarchy, adhocracy, and market (see Figure 1). For example, organizations exhibiting a clan culture tend to have an internal focus and a structure that emphasizes flexibility and change. Each of the four culture types is associated with certain assumptions, beliefs, values, behaviors, and effectiveness criteria (see Table 1). The clan culture, for example, puts special emphasis on values such as attachment, affiliation, or collaboration and is characterized by behaviors such as teamwork, participation, and employee involvement.

**FIGURE 1**
**Competing Values Framework**

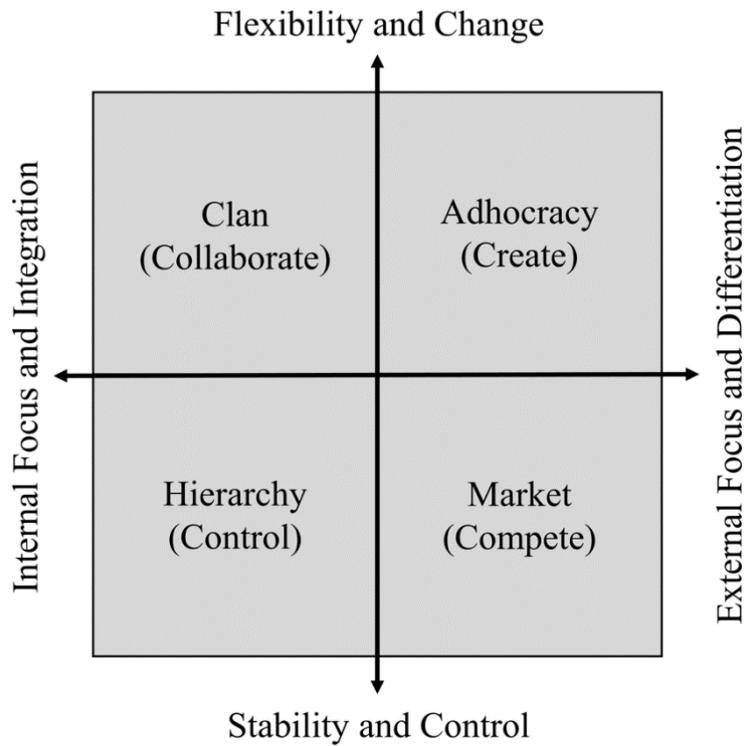

Adapted from Figure 3.1 in Cameron and Quinn (2011).

Importantly, the four culture types of the CVF are not mutually exclusive but tend to be positively correlated (Hartnell *et al.*, 2011). This means that a firm can exhibit several culture types at the same time. Therefore, some authors have suggested to refer to the culture types as culture dimensions (Hartnell *et al.*, 2019). We follow this notation in order to make clear that a firm being characterized by more than one type of culture is rather the norm than the exception.

**TABLE 1**
**CVF Culture Dimensions and Associated Attributes**

| Culture type | Assumptions | Beliefs | Values | Artifacts (behaviors) | Effectiveness criteria |
|---|---|---|---|---|---|
| **Clan** | Human affiliation | People behave appropriately when they have trust in, loyalty to, and membership in the organization. | Attachment, affiliation, collaboration, trust, and support | Teamwork, participation, employee involvement, and open communication | Employee satisfaction and commitment |
| **Adhocracy** | Change | People behave appropriately when they understand the importance and impact of the task. | Growth, stimulation, variety, autonomy, and attention to detail | Risk-taking, creativity, and adaptability | Innovation |
| **Market** | Achievement | People behave appropriately when they have clear objectives and are rewarded based on their achievements. | Communication, competition, competence, and achievement | Gathering customer and competitor information, goal-setting, planning, task focus, competitiveness, and aggressiveness | Increased market share, profit, product quality, and productivity |
| **Hierarchy** | Stability | People behave appropriately when they have clear roles and procedures are formally defined by rules and regulations. | Communication, routinization, formalization, and consistency | Conformity and predictability | Efficiency, timeliness, and smooth functioning |

Adapted from Table 13-1 in Quinn and Kimberly (1984) and Figure 2 in Hartnell *et al.* (2011).

## 2.2 Measuring Corporate Culture by Computational Linguistic Techniques

A common approach researchers have used to measure corporate culture is to administer a survey among members of the corporation. Starting from different theoretical backgrounds, different survey methods have been developed (see Chatman and O'Reilly (2016) for a review of the four most popular ones). For example, Cameron and Quinn (2011) proposed the Organizational Culture Assessment Instrument — a questionnaire to measure the four culture dimensions of the CVF. A drawback of measuring corporate culture with surveys is that assessing a large sample of corporations is both time- and resource-intensive. With recent advancements in computing power and new textual data bases becoming available, management scholars are increasingly using computational linguistics as an alternative (Chatman and Choi, 2022).

The earliest studies applying computational linguistic techniques to measure corporate culture have used the dictionary method. The idea of this method is to create a so-called dictionary or master text that is composed of words that supposedly signal the cultural trait that is to be measured. The existence of a given cultural trait is then measured by the textual similarity between the corresponding dictionary and the text documents of interest, for example all employee reviews of a given firm. To construct the dictionary, different approaches have been applied. While Grennan (2019) uses the lexical database WordNet, Li *et al.* (2021) apply a word embedding model.

Another computational technique researchers have applied to measure corporate culture from text documents is probabilistic topic modeling. It is an *unsupervised* machine learning method that is used to analyze which themes are covered in a given text. Corritore *et al.* (2020), for example, apply *latent Dirichlet allocation* (LDA; Blei *et al.*, 2003) — the topic model most widely used in studies on corporate culture — to measure cultural heterogeneity both between and within employees from employee reviews. However, especially if corporate culture is not the main focus of the text documents to be analyzed, many of the themes a topic model uncovers may not be related to corporate culture at all. Therefore, topic modeling is not well suited for the use case we look at, which is to identify specific culture dimensions from employee reviews, in which employees talk about more than just corporate culture. We will come back to this point in the final discussion.

We contribute to the literature on the measurement of corporate culture by introducing a state-of-the-art *supervised* machine learning approach. Supervised machine learning methods differ from unsupervised methods, such as topic modeling, in that a labeled data set is used. With supervised learning, an algorithm is trained to predict these labels. Our supervised learning approach consists of fine-tuning transformer-based language models for corporate culture classification. First introduced with BERT (Devlin *et al.*, 2018), transformer-based-language models now represent the state-of-the-art in natural language processing. They allow to capture complex aspects of textual information by taking the surrounding context into consideration and have been found to outperform traditional machine learning approaches in most natural language processing tasks (González-Carvajal and Garrido-Merchán, 2020). In financial sentiment classification, for example, BERT has been found to outperform simple machine learning approaches, such as naïve Bayes or support vector machine, as well as other deep learning algorithms, including convolutional neural networks and long short-term memory (Huang *et al.*, 2023).

# 3. Data

We collected 2,000 online employee reviews by drawing a random sample of all the reviews that were published on a leading employee review website between 2008 and 2018 and that matched the following two criteria: (i) The review was written by an employee or former employee working in the U.S and (ii) the employer is available in the Compustat database. We limit our sample to reviews of Compustat-listed firms to make our results comparable to other studies in which corporate culture is commonly linked to financial figures.

Each review includes several free text sections in which the reviewer can choose to talk about what she likes and dislikes about the employer and what changes she would recommend. We combined the free texts of the different sections in random order to obtain a single text for each review. All other information provided by the review besides the free texts was discarded. Next, we classified the reviews with respect to the employer's corporate culture. To do so, both authors and a research assistant (henceforth "labelers") independently went over all reviews. For each of the CVF's four culture dimensions, they assigned each review to one of three classes, depending on whether the review

- contains information in line with the culture dimension under question ("positive review"),
- contains information in conflict with the culture dimension under question ("negative review"), or
- does not allow any inference about the culture dimension under question ("neutral review").

In addition, the labelers assigned each review to exactly one of the four culture dimensions — the dimension that best fitted the overall tone of the review. This was done because although a firm can be characterized by several culture dimensions simultaneously, one of these dimensions usually dominates (Cameron and Quinn, 2011). In completing their labeling task, the labelers tried to stick as close as possible to the description of the four culture dimensions provided by Cameron and Quinn (2011) and Hartnell *et al.* (2011).

For each of the five classifications that needed to be made per review (one for each of the four culture dimensions and one for the dominant culture), the final classification was selected to be the majority vote of the three labelers. In case the labelers all had a different opinion, one of their decisions was picked at random. Table 2 provides the absolute frequency of review classifications by type of labeler agreement. All labelers agreeing on the same classification is the most frequent outcome for all five labeling tasks. The comparatively lower frequency of full agreement when picking the dominant culture can be explained by the fact that in line with the findings of Hartnell *et al.* (2011), reviews often include information that equally points to more than one culture dimension.

**TABLE 2**
**Number of Reviews by Type of Labeler Agreement**

| Type of agreement | Clan | Adhocracy | Market | Hierarchy | Dominant culture |
|---|---|---|---|---|---|
| No agreement | 61 | 17 | 56 | 34 | 176 |
| Two labelers agree | 850 | 458 | 760 | 632 | 890 |
| Full agreement | 1,089 | 1,525 | 1,184 | 1,334 | 934 |
| Sum | 2,000 | 2,000 | 2,000 | 2,000 | 2,000 |

We split our data set of 2,000 labeled employee reviews into a training (N=1,400), validation (N=200), and test set (N=400).

## 4. Model Fine-Tuning and Benchmarks
## 4.1 Transformer-Based Models

To build a state-of-the-art language model for corporate culture, we start-off with the two most widely-used language models that are based on the transformer architecture (Vaswani *et al.*, 2017): BERT (Devlin *et al.*, 2018) and RoBERTa (Liu *et al.*, 2019).[2] Both of these models have been trained on large unlabeled text corpora, including books and Wikipedia articles. As the name suggests, RoBERTa (*Robustly optimized BERT approach*) builds on BERT. However, it achieves higher performance in standard natural language processing tasks due to an improved training procedure and a much larger training data set (Liu *et al.*, 2019). For both BERT and RoBERTa, a base version with around 100 million parameters and a large version with around 350 million parameters exists. BERT and RoBERTa are general-purpose language models. In order to apply them on a specific natural language processing task, they need to be fine-tuned with the help of a labeled data set. This is exactly what we do in this paper. Using the 1,400 human-labeled employee reviews of our training set, we fine-tune the base and the large version of BERT and RoBERTa for the task of text classification in the domain of corporate culture. In the next step, we draw on the 200 labels of our validation set to evaluate different hyperparameters and to compare the resulting performance of the four models we look at. In line with Liu *et al.* (2019), we find that RoBERTa consistently outperforms BERT, both with respect to the base and the large version of the model. We therefore focus on RoBERTa in the following. Table 3 presents the hyperparameters of our final analysis.

**TABLE 3**
**Hyperparameters**

| | |
|---|---|
| Number of epochs | 8 |
| Weight decay | 0.01 |
| Learning rate | $1e^{-5}$ |
| Dropout rate | 0 |
| Batch size | 16 |
| Maximum sequence length | 200 |

## 4.2 TF-IDF-Based Text Classifiers

Since supervised machine learning has rarely been used to measure corporate culture, we benchmark our transformer-based models with other text classification methods that also use supervised learning but do not yet incorporate text embeddings as is the case with transformer-based models.[3] Such methods usually quantify the input text as a bag-of-word representation and apply machine learning algorithms to classify the text based on the occurrence and co-occurrences of these words (González-Carvajal and Garrido-Merchán, 2020). More specifically, texts are quantified with Term Frequency-Inverse Document Frequency (TF-IDF) matrices, which measure the occurrence of words in a text relative to the inverse number of occurrences in the entire document corpus. We follow González-Carvajal and Garrido-Merchán (2020) in computing the TF-IDF matrices with the TfidfVectorizer from sklearn. After the text has been transformed into a matrix representation, we apply traditional machine learning algorithms to classify the given text inputs. In particular, we use logistic regression, random forest, and XGBoost.

## 4.3 Dictionary Method

Following Grennan (2019) and Pasch (2018), we generate a dictionary for each of the CVF's culture dimensions, using the words that describe the associated assumptions, beliefs, values, behaviors, and effectiveness criteria that are mentioned in Quinn and Kimberly (1984). In addition, we included the synonyms and hyponyms of these words, using the WordNet library.

To measure the similarity between employee reviews and dictionaries, we compute for each employee review and each culture dimension the share of words that appear in the corresponding dictionary. Before doing so, we stemmed both the dictionary and the employee reviews ("create" and "creation" become "creat") and removed common stop words from the reviews (for example, "a", "and", "the"). Moreover, we take into account negations by subtracting the share of word stems of the dictionary that appear in a sentence with a negation word, such as "not" or "never". Online Appendix A presents the stemmed dictionaries we use.

The question of interest is how the dictionary method compares to our transformer-based language models in reproducing the human-generated corporate culture classification of employee reviews. For this purpose, the continuous similarity scores derived from the dictionary method need to be transformed into the discrete three-class-classification (positive, negative, or neutral) or four-class-classification (clan, adhocracy, hierarchy, market) of our labeling. We set the thresholds such that the relative frequency of the three or four classes matches the distribution in our training data set. For example, 21% of the reviews were labeled as showing evidence in accordance with a clan culture. Hence, the 21% of the reviews exhibiting the highest similarity with the clan dictionary were assigned to be positive with respect to a clan culture. We set the thresholds in this way to make sure that the dictionary method makes use of the same information as our language models. Our language models potentially picked up the distribution of reviews across labeling categories in the process of fine-tuning.

# 5. Results

## 5.1 Accuracy Scores

Table 4 presents accuracy scores, that is, the fraction of the 400 employee reviews of our test data set that were classified in line with human labelers, comparing different methods of text classification. In the category *dominant culture*, we analyze which of the CVF's four culture dimensions best fits the review. Hence, we would expect that a random classifier achieves an accuracy of 0.25. With respect to the four individual culture dimensions, a review can be classified as positive, negative, or neutral. Therefore, a random classifier should achieve an accuracy of 0.33. Table 4 also lists the performance of a classifier that always chooses the label that occurs most frequently (*majority class*). In the category *dominant culture*, this is the culture dimension *market*; in the case of the individual culture dimensions, this is the label *neutral*.

**TABLE 4**
**Accuracy Scores of Different Methods of Text Classification**

| Method of text classification | | Dominant culture | Clan | Adhocracy | Market | Hierarchy |
|---|---|---|---|---|---|---|
| Random | | 0.25 | 0.33 | 0.33 | 0.33 | 0.33 |
| Majority class | | 0.38 | 0.41 | 0.80 | 0.68 | 0.64 |
| Dictionary method | | 0.33 | 0.43 | 0.72 | 0.57 | 0.62 |
| TF-IDF | + logistic reg. | 0.48 | 0.61 | 0.81 | 0.69 | 0.69 |
| | + random forest | 0.46 | 0.58 | 0.81 | 0.69 | 0.74 |
| | + XGBoost | 0.47 | 0.59 | 0.82 | 0.74 | 0.74 |
| RoBERTa | base | 0.54 | 0.64 | 0.85 | 0.74 | 0.75 |
| | **large** | **0.63** | **0.68** | **0.89** | **0.77** | **0.81** |

Table reports accuracy scores, that is, the fraction of the 400 employee reviews of the test data set that were classified in line with human labelers. In the category *dominant culture*, it is determined which of the CVF's four culture dimensions best fits the review. In the categories *clan*, *adhocracy*, *market*, and *hierarchy*, it is determined whether the review contains information in accordance with, in contradiction to, or does not contain any information that allows inference about the existence of the culture dimension under consideration.

We find that the dictionary method outperforms a random classifier in all categories. However, with the exception of the clan category, it fails to outperform a classifier that always predicts the majority class.

Turning to the TF-IDF-based classifiers, we find that they outperform both a simple majority classifier as well as the dictionary method in all categories. With respect to our transformer-based language models, in turn, we observe that both RoBERTa-base and RoBERTa-large outperform all benchmarks in all categories. Moreover, RoBERTa-large consistently shows higher accuracy scores than the base version of RoBERTa, which is why we will focus on RoBERTa-large in the following. Overall, RoBERTa-large outperforms the dictionary method by 17 to 30 percentage points. It also outperforms TF-IDF-based classifiers by three to fifteen percentage points. The conclusions are qualitatively unchanged when looking at $F_1$ instead of accuracy scores (see Table B1 in the Online Appendix).

## 5.2 Why Does RoBERTa-large Outperform the Dictionary Method?

As the dictionary method still represents the standard approach in studies measuring pre-specified dimensions of corporate culture from text documents, we take a closer look what makes it less accurate than our language models that are based on RoBERTa-large. We look at all reviews that RoBERTa-large classified in the same way as human evaluators but that were misclassified by the dictionary method. We limit our attention to reviews with less than 50 words, however. While it is generally difficult to explain the predictions of RoBERTa-large since they do not follow from a simple rule-based approach, this is less of a problem in short reviews. This is because in short reviews, there is often just a single word or phrase that provides evidence in favor of one classification or another.

We found three major reasons for why many of the reviews that are misclassified by the dictionary method are still correctly classified by RoBERTa-large. Table 5 provides the relative frequency of these reasons for each of our five language models along with some examples. Since a review can be misclassified for several reasons and the allocation of cases across reasons is therefore not clear-cut, it is more the magnitude than the exact number of the relative frequencies that is of interest.

With respect to the clan and hierarchy dimension, in the majority of reviews that are correctly classified by RoBERTa-large but misclassified by the dictionary method, there is an expression that provides evidence in accordance with or in opposition to the culture dimension under consideration but does not use any of the word stems included in the respective dictionary. Hence, although actually positive or negative, the review is nevertheless classified as neutral when applying the dictionary method. For example, while the statement "they are quick to throw you under the bus" contradicts a clan culture, none of the statement's word stems is part of the dictionary for the clan dimension.

The second reason for why some of the reviews that are misclassified by the dictionary method are correctly classified by RoBERTa-large is that the review may use words that are included in the dictionary but in a different context. In this case, the review is wrongly classified as positive or negative when the dictionary methods is applied. In contrast, the review is correctly classified as neutral by RoBERTa-large since transformer-based language models take the surrounding context into account. For example, the remark that the job offers "a competitive compensation" does not signal a competitive corporate culture although the word "competitive" is used.

A closer inspection reveals that a significant number of misclassifications is caused by a limited set of word stems. Among these are the word stems "benefit" in the market dictionary and "time" in the hierarchy dictionary as they are regularly used in a context completely independent of corporate culture. Moreover, the word stems "advanc", "develop", "growth", and "grow" of the adhocracy dictionary turn out to be problematic as reviewers usually use them not to refer to an adhocracy culture but to talk about career advancement and personal development. Even removing these word stems from the dictionaries, however, does not significantly improve the accuracy of the dictionary method. This is because it also alters the threshold for the similarity between dictionary and employee review above or below which a review is classified as positive or negative, respectively. Hence, several reviews that so far were correctly classified as neutral now are misclassified as positive or negative due to other words included in the dictionary but used in the review in a different context. As selectively removing some word stems from our dictionary could be seen as arbitrary, we also reapply the dictionary method using the dictionaries that Fiordelisi and Ricci (2014) propose for measuring the CVF's four culture dimensions. Their dictionaries have been used in several other studies, including Nguyen *et al.* (2019) and Wang *et al.* (2021). Although their dictionaries do not contain any of the problematic word stems mentioned above, however, the accuracy scores we obtain when using these dictionaries are highly similar to those reported for the dictionary method in Table 4 (see Table C1 in the Online Appendix). The only exception is the clan dimension, for which we obtain a ten percentage point *lower* accuracy score when using their dictionary as opposed to ours. This is most probably because their dictionary for the clan dimension includes some very generic word stems, such as "employ" or "partner", which often are used in a different context than corporate culture.

The third reason for why RoBERTa-large exhibits a higher accuracy than the dictionary method is its superior ability in correctly interpreting the meaning of phrases that even in the absence of a negation have the opposite meaning of what the words by itself suggest. One example is the usage of the imperative. The statement "communicate more with your employees" signals a lack of communication and hence provides evidence in opposition to a clan culture although the word "communication" is used. As another example, RoBERTa-large understands that a firm that is "very slow in change" exhibits the opposite of an adhocracy culture although the word "change" appears.

In the case of the four language models for an individual culture dimension, less than ten percent of the reviews could not be assigned to any of the three reasons mentioned above. Among these are reviews for which more than one classification seemed reasonable. Hence, we keep these reviews separate from the other cases in which the misclassification clearly resulted from an inability to interpret the text correctly. The share of reviews that could not be assigned to any of the three reasons is higher in the case of the language model that predicts the dominant culture. This is because the variable dominant culture always has to take one of the four available dimensions. Therefore, some of the reviews that do not provide any conclusive information on corporate culture are correctly classified by RoBERTa but misclassified by the dictionary method just by chance.

**TABLE 5**
**Reasons Why Reviews Correctly Classified by RoBERTa-large Were Misclassified by Dictionary Method**

| Culture dimension | No. of reviews | Reason why review was misclassified by dictionary method | Share of reviews | Example | Classification of example RoBERTa-large | Classification of example dictionary method |
|---|---|---|---|---|---|---|
| Clan | 62 | Wording without words from dictionary | 81% | They are quick to throw you under the bus. | negative | neutral |
| | | Words from dictionary in different context | 8% | It is fun **help**ing customers find the right product. | neutral | positive |
| | | Did not capture opposite meaning | 10% | **Commun**icate more with your employees. | negative | positive |
| | | Other | 2% | | | |
| Adhocracy | 54 | Wording without words from dictionary | 19% | No room for new technologies. | negative | neutral |
| | | Words from dictionary in different context | 69% | All the red tape **creat**es excessive work. | neutral | positive |
| | | Did not capture opposite meaning | 4% | Very slow in **chang**e. | negative | positive |
| | | Other | 9% | | | |
| Market | 65 | Wording without words from dictionary | 22% | You are nothing but a number. | positive | neutral |
| | | Words from dictionary in different context | 71% | Offers **compet**itive compensation. | neutral | positive |
| | | Did not capture opposite meaning | 2% | Reward actual **achiev**ement, not face time. | negative | positive |
| | | Other | 6% | | | |
| Hierarchy | 54 | Wording without words from dictionary | 52% | There is too much change. | negative | neutral |
| | | Words from dictionary in different context | 41% | Company **consist**s of nothing but thieves. | neutral | positive |
| | | Did not capture opposite meaning | 6% | The firm is constantly adjusting its **structur**e. | negative | positive |
| | | Other | 2% | | | |
| Dominant culture | 83 | Wording without words from dictionary | 63% | Very cumbersome and antiquated processes. | hierarchy | random |
| | | Words from dictionary in different context | 11% | Corporate **object**ives shift constantly. | adhocracy | market |
| | | Did not capture opposite meaning | 0% | **Innov**ation takes forever. | hierarchy | adhocracy |
| | | Other | 27% | | | |

Table considers all reviews with less than 50 words that RoBERTa-large classified in the same way as human labelers but that were misclassified by the dictionary method. In the example sentences, the word stems that are part of the dictionary of the respective culture dimension are marked in bold.

# 6. Discussion

We fine-tune transformer-based language models for classifying employee reviews with respect to their alignment with the four culture dimensions of the CVF. In principal, our analysis could be repeated using a different framework than the CVF. This would require creating a new data set of employee reviews labeled with respect to a different set of culture dimensions. Due to their advantages in understanding the meaning of words and taking semantic context into account, we expect that transformer-based language models would similarly outperform other approaches of text classification even if a different set of culture dimensions was used.

We evaluate the performance of our transformer-based language models by comparing them to the dictionary method as well as supervised machine learning approaches based on TF-IDF matrices. We do not apply topic modeling, however, since it is not well suited for identifying specific culture dimensions from text documents that are not specifically related to corporate culture. Since the employee reviews we use are free texts, they contain information on various job-related issues, many of which are not related to corporate culture at all. Not surprisingly, therefore, applying a topic model to employee reviews predominantly yields groupings of words that are unrelated to corporate culture. Hence, although using a topic model does not require to specify ex ante what cultural content employee reviews are assumed to cover, it is nevertheless associated with making subjective decisions, for example, what part of the text to consider or how to interpret, label, and filter the topics obtained (Schmiedel *et al.*, 2019).

We are aware of the fact that the accuracy of the computational linguistic techniques we evaluate may depend on the labeled data set we propose. The superior accuracy of the language models for corporate culture we propose, however, is well in line with what has been found about the performance of transformer-based language models on natural processing tasks more generally (González-Carvajal and Garrido-Merchán, 2020). By making our models publicly available, we allow other researchers to verify and to make use of the superior performance of transformer-based language models in the domain of corporate culture, using their own data.

A common drawback of language models that are based on machine learning is that it is difficult to explain the predictions they deliver. In this regard, our language models are no exception. While we focus on showing that transformer-based language models are more accurate in measuring corporate culture from text documents than traditional approaches of text classification, we also provided some insights on the sources of this higher degree of accuracy. Huang *et al.* (2023) follow a different approach to explore the superior classification performance of transformer-based language models. They analyze how the accuracy changes when the words in the observations to be labeled are randomized. They observe that this leads to a sizeable drop in the accuracy of their transformer-based models but decreases the accuracy of other machine learning algorithms only moderately, suggesting that transformer-based language models much better capture contextual information.

      Future work should explore more thoroughly what are the advantages of transformer-based language models in measuring corporate culture or other organizational phenomena but should also look at their limitations. This requires providing an explanation for the models' predictions, for example by using a local interpretable model (Ribeiro *et al.*, 2016).

# Footnotes

[1] Our language models are uploaded under https://huggingface.co/CultureBERT. In addition, we provide a tutorial on how to apply our models to measure corporate culture from text documents under https://github.com/Stefan-Pasch/CultureBERT.

[2] On the Huggingface model hub, as of November 2022, BERT and RoBERTa show the highest number of downloads among the English language models used for text classification. See https://huggingface.co/models.

[3] In fact, we are not aware of any application of supervised machine learning to classify text with respect to the underlying corporate culture dimensions.

# Supplementary Online Appendix

## Online Appendix A: Dictionaries

**TABLE A1**
**Stemmed Dictionaries**

| Culture dimension | | Stemmed dictionary |
|---|---|---|
| Clan | positive | clan, affili, connect, attach, bond, adher, adhes, trust, collabor, coaction, collaboration, quisling, join_forc, cooper, togeth, get_togeth, play_along, go_along, support, help, assisten, encourag, teamwork, engag, particip, involv, involut, commit, open, recept, overt, commun, satisfact, gratif, comfort, pride, allegi, loyalti, dedic, devot, not_uncoop, not_disjunct, not_unaccommod, not_discourag, not_unsupport, not_disconnect, not_distrust, not_not-engag, not_nonparticip, not_non-involv, not_unrecept, not_covert, not_dissatisfact, not_humil, not_discomfort, not_disloyalti |
| | negative | uncoop, disjunct, unaccommod, discourag, unsupport, disconnect, distrust, non-engag, nonparticip, non-involv, unrecept, covert, dissatisfact, humil, discomfort, disloyalti, not_clan, not_affili, not_connect, not_attach, not_bond, not_adher, not_adhes, not_trust, not_collabor, not_coaction, not_collaboration, not_quisling, not_join_forc, not_cooper, not_togeth, not_get_togeth, not_play_along, not_go_along, not_support, not_help, not_assisten, not_encourag, not_teamwork, not_engag, not_particip, not_involv, not_involut, not_commit, not_open, not_recept, not_overt, not_commun, not_satisfact, not_gratif, not_comfort, not_pride, not_allegi, not_loyalti, not_dedic, not_devot |
| Adhocracy | positive | hazard, jeopardi, peril, risk, endanger, danger, gambl, adventur, jeopard, adapt, flexibl, pliabil, plianci, pliant, advanc, forward-look, innov, modern, groundbreak, invent, excogit, concept, initi, found, foundat, creation, introduct, instaur, attent, detail, technic, high_spot, creat, produc, alter, modif, chang, varieti, modifi, growth, grow, matur, develop, mergenc, outgrowth, stimul, divers, multifari, creativ, creative_think, adhocraci, not_safeti, not_safe, not_unadapt, not_inflex, not_intract, not_regress, not_demot, not_old_styl, not_nonmodern, not_abolish, not_misconcept, not_inflect, not_imprecis, not_inaccur, not_inexact, not_inattent, not_nondevelop, not_uncr |
| | negative | safeti, safe, unadapt, inflex, intract, regress, demot, old_styl, nonmodern, abolish, misconcept, inflect, imprecis, inaccur, inexact, inattent, nondevelop, uncr, not_hazard, not_jeopardi, not_peril, not_risk, not_endanger, not_danger, not_gambl, not_adventur, not_jeopard, not_adapt, not_flexibl, not_pliabil, not_plianci, not_pliant, not_advanc, not_forward-look, not_innov, not_modern, not_groundbreak, not_invent, not_excogit, not_concept, not_initi, not_found, not_foundat, not_creation, not_introduct, not_instaur, not_attent, not_detail, not_technic, not_high_spot, not_creat, not_produc, not_alter, not_modif, not_chang, not_varieti, not_modifi, not_growth, not_grow, not_matur, not_develop, not_mergenc, not_outgrowth, not_stimul, not_divers, not_multifari, not_creativ, not_creative_think, not_adhocraci |
| Market | positive | market, commerci, commercialis, offer, contest, competit, rivalri, accomplish, achiev, commun, compet, goal, destin, aim, object, target, intent, purpos, plan, prepar, provis, be_aft, project, contriv, design, task, aggress, belliger, pugnac, fast-grow, strong-grow, profit, earn, gain, benefit, qualiti, product, gener, fertil, not_cooper, not_incompet, not_subject, nott_resolut, not_unaggress, not_noncompetit, not_unproduc |
| | negative | cooper, incompet, subject, resolut, unaggress, noncompetit, unproduct, not_market, not_commerci, not_commercialis, not_offer, not_contest, not_competit, not_rivalri, not_accomplish, not_achiev, not_commun, not_compet, not_goal, not_destin, not_aim, not_object, not_target, not_intent, not_purpos, not_plan, not_prepar, not_provis, not_be_aft, not_project, not_contriv, not_design, not_task, not_aggress, not_belliger, not_pugnac, not_fast-grow, not_strong-grow, not_profit, not_earn, not_gain, not_benefit, not_qualiti, not_product, not_gener, not_fertil |
| Hierarchy | positive | hierarchi, structur, hierarch, control, restraint, command, stabl, stabil, constanc, static, unchang, procedur, function, routin, formal, formalis, convent, schemat, consist, reproduc, coher, logic, order, uniform, conform, complianc, abid, accord, predict, effici, effect, time, season, smooth, not_nonhierarch, not_unrestraint, not_instabl, not_unstabl, not_inconst, not_malfunct, not_inform, not_unconvent, not_inconsist, not_incoher, not_illog, not_unseason, not_ineffect, not_ineffici, not_unpredict, not_nonconform, not_noncompli |
| | negative | nonhierarch, unrestraint, instabl, unstabl, inconst, malfunct, inform, unconvent, inconsist, incoher, illog, unseason, ineffect, ineffici, unpredict, nonconform, noncompli, not_hierarchi, not_structur, not_hierarch, not_control, not_restraint, not_command, not_stabl, not_stabil, not_constanc, not_static, not_unchang, not_procedur, not_function, not_routin, not_formal, not_formalis, not_convent, not_schemat, not_consist, not_reproduc, not_coher, not_logic, not_order, not_uniform, not_conform, not_complianc, not_abid, not_accord, not_predict, not_effici, not_effect, not_time, not_season, not_smooth |

Table presents the stemmed dictionaries we use to apply the dictionary method as described in section 4.3. We compute for each employee review and each culture dimension the share of words that appear in the corresponding dictionary of *positive* word stems. To take into account negations, we subtract from this share the share of words that appear in the corresponding dictionary of *negative* word stems. A word stem that appears in the dictionary with the prefix "not_" is counted only if it appears in a sentence with a negation word, such as "not" or "never".

# Online Appendix B: $F_1$ Scores

**TABLE B1**
**$F_1$ Scores of Different Methods of Text Classification**

| Method of text classification | | Dominant culture | Clan | Adhocracy | Market | Hierarchy |
|---|---|---|---|---|---|---|
| Random | | 0.25 | 0.33 | 0.33 | 0.33 | 0.33 |
| Majority class | | 0.20 | 0.24 | 0.71 | 0.55 | 0.51 |
| Dictionary method | | 0.33 | 0.36 | 0.70 | 0.54 | 0.55 |
| TF-IDF | + logistic reg. | 0.44 | 0.60 | 0.73 | 0.60 | 0.62 |
| | + random forest | 0.41 | 0.55 | 0.74 | 0.59 | 0.70 |
| | + XGBoost | 0.44 | 0.59 | 0.77 | 0.70 | 0.72 |
| RoBERTa | base | 0.54 | 0.63 | 0.84 | 0.72 | 0.74 |
| | **large** | **0.63** | **0.69** | **0.88** | **0.76** | **0.81** |

Table reports $F_1$ scores obtained when classifying the 400 employee reviews of the test data set. See notes of Table 4.

# Online Appendix C: Results Applying Dictionaries from Fiordelisi and Ricci (2014)

**TABLE C1**
**Accuracy and $F_1$ Scores Using Dictionaries from Fiordelisi and Ricci (2014)**

| Method of text classification | Score | Dominant culture | Clan | Adhocracy | Market | Hierarchy |
|---|---|---|---|---|---|---|
| Dictionary method | Accuracy | 0.32 | 0.34 | 0.75 | 0.58 | 0.62 |
| + word count | $F_1$ | 0.34 | 0.34 | 0.72 | 0.58 | 0.54 |

Table presents accuracy scores and $F_1$ scores that are obtained applying the dictionary method as described in section 4.3 but using the dictionaries proposed by Fiordelisi and Ricci (2014). See notes of Table 4.